\documentclass[11pt]{article}
\usepackage{acl2014}
\usepackage[utf8]{inputenc}
\usepackage{times}
\usepackage{url}
\usepackage{latexsym}
\usepackage{amsmath}
\usepackage{paralist}
\usepackage{pdfpages}
\DeclareMathSizes{11}{9}{13}{9}

\title{Supertagging: Introduction, learning, and application}
\author{Taraka Rama K.\\
Språkbanken\\
University of Göteborg}

\date{}

\begin{document}
\maketitle
\begin{abstract}
Supertagging is an 
approach originally developed by \newcite{bangalore1999supertagging} to 
improve the parsing efficiency. In the beginning, the scholars used small 
training datasets 
and somewhat naïve smoothing techniques to learn the probability distributions 
of supertags. Since its inception, the applicability of Supertags has 
been explored for TAG (tree-adjoining grammar) formalism as well as other 
related yet, different formalisms such as CCG. This article 
will try to summarize the various chapters, relevant to statistical parsing, 
from the most recent edited book volume~\cite{bangalore2010supertagging}. The 
chapters were selected so as to blend the learning of supertags, its 
integration into full-scale parsing, and in semantic parsing.
\end{abstract}

\section{Introduction}\label{sec:intro}
\footnote{Submitted as a term paper for Statistical 
Parsing course given by Prof. Dr. Joakim Nivre.}The main theme of supertagging 
is \emph{complicate locally, simplify 
globally} (CLSG). Originally proposed for LTAG (Lexicalized Tree-Adjoining 
Grammars) framework, supertagging aims at reducing the overall parsing 
complexity by disambiguating the supertags of lexical units locally which 
was espoused in \newcite{bangalore1999supertagging}. The task of supertagging 
is likened to the task of POS tagging which is a well-defined and 
well-explored task in computational linguistics. In this section, I will give a 
brief introduction to LTAG and attempt to couch supertagging in the LTAG 
framework.

CFG is a string re-write formalism whereas, LTAG is a tree 
re-write formalism. LTAG comes with two kinds of elementary trees: 
\emph{initial} ($\beta$) and \emph{auxiliary} ($\alpha$). There are two kinds 
of operations namely, substitution and adjoining operation. The parse tree for 
a sentence is derived through a sequence of operations applied to the 
elementary trees anchored in the lexical items.

A $\beta$ tree can be substituted at its \emph{frontiers} 
whereas a $\alpha$ tree undergoes adjoining operation. A $\alpha$ tree is a 
elementary tree which has a root node $X$ and a non-terminal node $X^*$. The 
non-terminal node $X^*$ can undergo adjoining operation. An adjoining operation 
involves detaching the tree at node $X$ and splicing a $\alpha$ tree at the 
node $X$. In the next step, the detached tree is substituted at the node marked 
with $^*$.  As mentioned earlier, each elementary tree is lexicalized in LTAG.

A substitution operation works towards growing a tree at the leaves and cannot 
operate at internal nodes whereas, a adjoining operation works at the tree 
internal nodes and grows the tree such that local dependencies encoded in the 
elementary trees can encode long distance dependencies. Object extraction 
construction is an example where adjoining operation increases the distance 
between locally dependent lexical units in an initial tree.

LTAG works, primarily, with elementary trees and combines the elementary trees 
through substitution and adjoining operations to derive a sentence. If the 
correct elementary tree of each lexical item in a sentence is determined 
accurately beforehand, then a parser can be employed to derive the complete 
parse tree with relative ease by applying the adjoining and substitution 
operations. 
The final sequence of disambiguated elementary parses delivered by a supertagger 
is referred to as \emph{almost parse} by the authors. The supertags for a given 
language is obtained by parsing a corpus through a hand-written LTAG grammar. 
Another method is to use a pre-annotated tree bank such as Penn 
treebank~\cite{marcus1993building} to extract the set of supertags to train a 
supertagger.

The original experiments \cite{joshi1994disambiguation}, to the test the 
efficacy of supertagging, were conducted on a Wall Street Journal Corpus (WSJ). 
The training corpus consisted of 1 million words and the test corpus consisted 
of $47,000$ words. They find that the supertag ambiguity is about $15$ to $20$ 
supertags per word. In these experiments, the baseline is quite 
straightforward. They assign the most 
frequent supertag as the supertag for a word. The baseline system achieves an 
accuracy of $75\%$ which means that $75\%$ of the words are given the right 
supertag. Then, they train a trigram-based POS tagger on their training corpus 
and 
achieve an impressive accuracy of $92\%$. This work forms the foundation for 
the work surveyed in this article.

The rest of this paper is organized as follows. 
Section~\ref{sec:suptagpars} focuses on the work which 
learns \cite{shendiscriminative} combines parsing with 
supertagging \cite{sarkar2007combining}. In tandem with the organization of 
the book, the section~\ref{sec:relforms} surveys the place of 
supertagging in other grammar formalisms such 
as: CCG \cite{clark2010supertagging} and PCFG parsing with latent 
annotations~\cite{matsuzakiprobabilistic}.
Section \ref{sec:analapps} surveys the application of supertagging to semantic 
role labeling~\cite{chensemantic}. Finally, we conclude the paper in 
section~\ref{sec:stsc} by summarizing few statistical parsing articles related 
to supertagging (since 2010).

\section{Extraction and Learning of Supertags}\label{sec:suptagpars}
\subsection{Supertagging and Parsing Efficiency}
\newcite{sarkar2007combining} attempts to combine supertagging with full-scale 
LTAG parsing. There are at least two arguments to claim that this idea can 
yield higher parsing accuracies. The first argument is motivated through parsing 
efficiency. Given a highly lexicalized grammar such as LTAG, the \emph{syntactic 
lexical ambiguity} and \emph{sentence complexity} might be the \emph{dominant} 
factors that affect parsing efficiency. If these two factors play a major role 
in improving or decreasing the parsing efficiency, it certainly means that 
supertagging -- which was originally designed to disambiguate locally -- can be 
used for first purpose. Also, a POS-style tagger is much less affected by a 
constraint such as sentence length. Secondly, Sarkar combines the supertagger 
with a full-scale LTAG parser in a co-training framework where two 
conditionally independent parsers supplement each other by starting from a 
small seed list of training examples to bootstrap a LTAG parser from a large 
unannotated corpus.

\newcite{sarkar2007combining} tests the motivations for the first argument by 
performing two sets of experiments varying two parameters: number of derivation 
trees and number of clauses, per sentence. The motivation for experimenting 
with these parameters comes from the worst case time and space complexity of the 
Earley-style chart parser used by \newcite{schabes1994left}. The worst case time 
complexity of this parser is in the order of $|A|\cdot |I \cup A|\cdot N \cdot 
n^6$ and the space complexity is in the order of $|I \cup A| \cdot N \cdot n^4$ 
where, $I$ is the set of initial trees, $A$ is the set of auxiliary trees, $N$ 
is the maximum number of nodes in an elementary trees, and $n$ is the length of 
the input string. Given this complexity, Sarkar tries to determine which 
parameter correlates the most with parsing efficiency.

The process of operations in a LTAG framework can be treated as attaching and 
rearranging elementary trees to a root node. Hence, the parsing task breaks 
down into two steps:
\begin{compactenum}
 \item Assign a probable set of elementary trees to a lexical item.
\item Find the correct attachments between these elementary trees to arrive at 
all the 
parses for a sentence.
\end{compactenum}
In a sentence, the number of elementary trees as well as the 
number of clauses might, as well, increase with the sentence length. Sarkar 
tests his 
claims by employing an automatically extracted LTAG treebank grammar and a 
chart-based head-corner parser. The test set consists of $2250$ sentences.  
Also, the parser produces all the parses for a given sentence in a packed 
forest representation.

Overall, Sarkar makes the following observations regarding the relation 
between parsing efficiency; and the number of trees and clauses:
\begin{compactenum}
 \item There is a correlation of $0.65$ between parse times and sentence length.
\item There is a stronger correlation of $0.82$ between the parse times and the 
number of trees in a sentence.
\item The correlation between number of deviations and 
the parse times is not very strong.
\item The number of clauses do not increase with sentence length. The parsing 
efficiency seems to be independent of the number of clauses in a sentence.
\item Finally, in an oracle experiment, when the parser is supplied with the 
gold-standard elementary trees, for an input string; the parsing times ($< 
1~sec.$) drop drastically.
\end{compactenum}
The oracle experiment shows that \emph{apriori} knowledge of the right 
elementary trees increases the parsing efficiency. This fact can be 
used to substantiate the claim whether supertagging helps parsing efficiency. 
Sarkar also tests if a n-best supertags per word are beneficial for reducing 
the parsing time. Therefore, he supplies $60$-best supertags per word as 
an input to the parser. Sarkar notes that the parsing time is reduced by more 
than $25$ times. In the co-training experiment, the author finds that using 
supertagging as the other model improved the labeled precision and recall from a 
baseline probabilistic parser. Overall, this work shows that supertagging is 
useful for both improving the a full-scale parser as well as bootstrapping the 
same parser in a data-paucity scenario.

\subsection{Learning supertags}
In this subsection, I summarize the work of \newcite{shendiscriminative} who 
attempts to improve the supertagging accuracies by modeling the problem in a 
sequence labeling framework that addresses the \emph{label bias} problem 
inherent in a HMM-style sequence tagging.

\newcite{shendiscriminative} introduces a new method to learn supertags. The 
author models supertag learning as a sequential learning problem. In doing so, 
he explores the local probability dependent models such as PMMs 
(Projection-based Markov models) and tries to pass their limitations by using a 
discriminative learning model such as SNoW (Sparse Network of Winnow; 
Roth 1998). Shen compares the performance of his method with that of 
\newcite{chen2001towards}. Further, he applies his system at the task of NP 
chunking and finds that use of supertags improves the standard NP chunker's 
performance.

Continuing the trend in the book, Chen explores the use of supertags to 
improve the performance of a baseline Transformational-based Learning NP 
chunker. The author begins his article by noting that the performance of NP 
chunker drops when supplied with supertags automatically generated by a trigram 
supertagger. The author hypothesises that automatically generated supertags 
provide more noise than information. The author reviews the previous work in 
data-driven supertagging scenario~\cite{srinivas1997performance,chen1999new} 
and finds that a combination of lexicalized and supertag head based contextual 
features (\emph{mixed} model) coupled with few heuristic rules improves the 
supertagger accuracy. Further, \newcite{chen1999new} employed a pair-wise voting 
scheme to combine their models that yielded the highest accuracy. The accuracy 
of the NP chunking system increased when they used the automatically generated 
supertags as an input. And, they employed heuristic rules to identify the NP 
chunks. However, as Shen notes, it is not clear if this high accuracy is due to 
the supertags or due to the heuristic rules.

Shen models the task of supertagging learning as a sequential learning problem. 
He begins by observing that discriminative learning of supertagging can be 
achieved by training a classifier for the corresponding lexical item's POS tag. 
Although a joint learning of POS tags and supertags can be a feasible approach, 
the author prefers a two-step approach where, in the first step, a Brill POS 
tagger is used to tag the training and testing datasets. The supertag modeling 
is done as followed. Let $W = w_1\dots w_n$ be a sentence, $Q=q_1\dots q_n$ be 
the POS tags, and $T=t_1\dots t_n$ be the corresponding supertags. Then the 
probability of assignment of a supertag $t_i$ is modeled as:
\begin{equation}
 P(t_i|t_{1\dots i-1}, W, Q) \equiv P_{q_i}(t_i|t_{1\dots i-1}, W, Q)
\end{equation}
Hence, a separate classifier is built for each POS tag. This approach has the 
following advantages:
\begin{compactenum}
 \item Data sparsity problem: There are more than $400$ supertags as compared 
to the number of POS-tags ($<100$).
\item The classifier can focus on supertags belonging to each POS tag and learn 
the difficult cases.
\end{compactenum}
The author employs a five word window plus two previous head supertags as 
features to train the classifier. The system is bidirectional: scans 
the sentences in both the directions; and employs a pairwise voting to supertag 
a sentence. The author tests his system on section 20 of WSJ corpus. The author 
notes that his supertagger achieves an error reduction of $13\%$ on 
automatically extracted LTAG grammar based supertags. The SNoW based supertagger 
beats \newcite{chen2001towards} with a difference of $0.59\%$ in terms of 
accuracy.

Having established that his supertagging system beats the previously 
established best systems, Shen proceeds to test its efficiency in NP chunking. 
In NP chunking experiments, sections 2--14 and 21--24 are used to train the 
supertagger. Section 19 is used to tune the supertagger and section 20 for 
testing purposes. The author finds that SNoW system's supertags helped improve 
the overall accuracy of IOB tagging as well as the precision, recall and, 
F-score of the NP chunker. The author also performs two oracle experiments by 
supplying gold standard POS tags and supertags to the NP chunker only to find 
that the NP chunker performs much better across all the domains. Thus, he 
concludes that improving the accuracy of supertagging can only benefit NP 
chunking task.

\section{Supertags in related formalisms}\label{sec:relforms}
\subsection{Combinatory Categorial Grammar}
Parsing in Combinatory  Categorial Grammar, referred to as CCG 
\cite{steedman2000syntactic}, is quite similar to LTAG and consists of two 
steps. The first step consists of assigning elementary syntactic structures 
to 
the words in a sentence; the second step consists of combining the elementary 
structures to derive the full parse tree. In a similar spirit, 
\newcite{clark2002supertagging} and \newcite{clark2004importance} introduce and 
establish supertagging in the CCG framework. This line of work is continued in 
\newcite{clark2010supertagging} where, the authors develop a multi-tagger based 
on the Maximum-Entropy based supertagger. The authors also test the importance 
of a \emph{tag-dictionary} -- a dictionary that maps supertags to words -- and 
the $n$-best lists (per word) at the task of CCG parser 
efficiency.

\newcite{clark2004importance} develop a supertagger based on an automatically 
extracted CCG grammar from Penn Treebank. The authors note that the CCG 
parser's parsing time reduces when the correct supertags are supplied to the 
parser. In an analogy with a zig-saw puzzle \cite{bangalore2010supertagging}, 
the parser is initially supplied with a small number of supertags, and the 
parser receives more supertags only if the parser fails to derive a full parse 
tree. The authors test the performance of supertagging and its importance to 
parsing in relation to the following three parameters:
\begin{compactenum}
 \item The size of tag-dictionary.
\item The purity of POS tags.
\item Multi-tags for each word within an admissible range of variation 
(quantified in terms of a bestness parameter $\beta$).
\end{compactenum}

The previous work in CCG supertagging \cite{clark2002supertagging} reports a 
higher accuracy in comparison to the LTAG supertagging work. The authors attempt 
to explain this difference in the following terms:
\begin{compactenum}
\item The formalisms treat syntactic constructions differently.
\item The CCG grammar extraction process caused the CCG supertag set to be much 
smaller in comparison with the LTAG supertag set. 
\end{compactenum}

The authors note that by restricting the size of original tag-dictionary 
through a frequency cut-off (at least 10), the overall size of the supertag set 
is reduced three-fold. This step contributes to decreasing the supertag 
ambiguity and also speeding the parser. The authors employ a Maximum-Entropy 
(MaxEnt) framework to assign the right supertag to a word. The contextual 
features for training their MaxEnt framework are simple; and are based on word 
features extracted from a window-size of five, a combination of unigram and 
bigram POS tags, 
and lexical categories (supertags) of the previous two words.

Clark and Curran use a one-best category per word and gold standard POS tags to 
train their supertagger and find that the word accuracy is $92.6\%$ and the 
sentence accuracy is $36.8\%$. However, automatically assigned POS tags 
decrease the word and sentence accuracies to $91.5\%$ and $32.7\%$ 
respectively. Hence, the scholars proceed to test if ambiguity in the 
assignment of categories per word influences the word and sentence accuracies 
of the supertagger. The number of lexical categories per word is computed using 
a $n$-best list of lexical categories computed using a forward-backward 
algorithm. By 
allowing a large latitude in the range of $\beta$, the supertagger performs 
almost as well as the supertagger model trained on gold standard POS-tags. The 
parse times of the parser reduce drastically when the parser is tested in 
combination with the supertagger. In conclusion, the CCG parser assigns parses 
to $99.6\%$ of the sentences as compared to the LTAG parser which assigns 
complete parses to $60\%$ of the sentences.

\subsection{PCFG with latent annotations}
The chapter by \newcite{matsuzakiprobabilistic} is not directly related to 
supertagging but bears on the automatic annotation of ancestor nodes in a 
classic PCFG parsing setting. The authors attempt to weaken the strong 
independence assumptions made by a PCFG parser by annotating each non-terminal 
node in a parse tree with a latent variable. This approach has been tried 
previously for PCFG parsing in different contexts.\footnote{See 
\newcite[247]{nivre2010statistical} for a more complete list of references.} 
The novelty of the idea presented in this book chapter is that the non-terminal 
ancestor nodes in a parse tree are labeled with a latent variable.

\newcite{matsuzakiprobabilistic} describe and develop the formulas for the 
estimation of the rule probabilities in an EM-style algorithm. The authors 
note that the task of finding the best parse tree is an NP-hard problem. 
Subsequently, the authors use three different approximation techniques to 
reduce the tree space. The first approach consists of generating $n$-best parse 
trees using a PCFG model and then selecting the best parse tree based on 
the PCFG-LA model. The second approximation approach consists of using a 
viterbi approximation to the standard PCFG model with latent annotations. The 
third approximation consists of approximating the original objective function 
with weak independence assumptions.

The authors test the different 
approximations with different parse tree binarization techniques. The nature of 
binarization does not effect the accuracy of the three approximations. Finally, 
the authors compare their approach with state-of-the-art lexicalized parsers 
and note that their parser is comparable to the performance of 
unlexicalized parsers~\cite{klein2003accurate} but does not come close to the 
lexicalized parsers such as Charniak~\cite{Charniak:00}.

\section{Application to Semantic Labeling}\label{sec:analapps}
\newcite{chensemantic} tests the efficiency of supertagging coupled with two 
different kinds of parsers (Light-weight dependency Parser 
[LDA]\footnote{Originally developed by \newcite{bangalore1999supertagging}, 
LDA is a fast quadratic-time parser which takes a supertagged sentence as 
input and constructs a dependency tree by finding the local syntactic heads 
and then, links each argument to its predicate.} and a probabilistic LTAG 
parser \cite{schabes1992stochastic}) for the task of semantic labeling. This 
chapter is a continuation of the work done by \newcite{chen2003use} on a similar 
task. As noted by Chen, the earlier work used a pre-released version of 
PropBank \cite{palmer2005proposition} for the same task. The current book 
chapter focuses on three issues:
\begin{compactenum}
 \item Use of deep linguistic features improves the performance of semantic 
labeler.
\item A LDA parser can perform as well as a full LTAG parser at the task of 
semantic parsing.
\item A unified syntactic and semantic TAG parser is preferable to a pipelined 
TAG parser and semantic labeler.
\end{compactenum}

The author limits the task of semantic labeling to the identification 
of predicate argument labels. The author refrains from roleset labeling as well 
as adjunct labeling since he notes that even a majority class label assigner 
achieves an accuracy of $88.3\%$ on the task of roleset labeling.

The PropBank is labeled on the top of PTB which is useful in extracting surface 
syntactic features for the task of semantic labeling. In an earlier work, 
\newcite{gildea2002necessity} use surface syntactic features such as Head word, 
Phrase type, Path etc., for training their discriminative classifier-based 
system. These features are 
readily extracted from PropBank due to its origins. Except for \emph{Voice}, 
rest of the features are surface syntactic features.

Chen hypothesizes that deep linguistic features may be useful for the task of 
semantic labeling. The author's methodology can be summarized as follows:
\begin{compactenum}
 \item Generate features for various levels of linguistic analysis. This is 
done by extracting different kinds of TAGs from PropBank annotations.
\item Use of the extracted TAGs for the prediction of semantic roles given: 
gold parse trees and raw text parsed using LDA and LTAG parser.
\end{compactenum}

For the first task, the author extracts two kinds of TAGs: SYNT-TAG 
(Syntactic TAG) and SEM-TAG (Semantic TAG). SYNT-TAG has two kinds of features: 
\emph{surface} and \emph{deep} syntactic features. The former consists of 
features which are extracted from hand-crafted rules based on the argument's 
position. The second features consist of extracting the \emph{trace} of a 
transformation that yields a construction. SEM-TAG consists of SYNT-TAG's 
features complemented with the PropBank argument labels. The author also uses 
supertag based features modeled in the lines of 
\newcite{gildea2002necessity}\footnote{Henceforth, referred to as GP.}. The 
author finds that \emph{surface}, \emph{deep}, and \emph{supertag} features beat 
the GP features at the task of semantic labeling on gold standard parses.

As a prerequisite to parsing the raw text, the author tests the performance of 
the standard supertagger \cite{bangalore1999supertagging} on SYNT-TAG and 
SEM-TAG for various sizes of n-best list supertags. The author finds that 
as $n\to \infty$ the supertaggers based on both kinds of TAGs achieve high 
accuracies of $97\%$. The author then proceeds to test the performance of 
SYNT-TAG and SEM-TAG for the task of semantic argument recognition using 
supertagged raw text coupled with LDA or a LTAG parser. The performance of 
the supertagger is evaluated on boundary detection and argument's headword 
detection. Overall, the system's performance is best on the task of 
boundary detection when trained on SEM-TAG grammar coupled with full 
statistical parser. The system's performance is not hurt when the same system 
is evaluated on headword detection task. The author concludes his article by 
claiming that \emph{deep} features in combination with a supertagged raw text 
and a statistical parser trained on SEM-TAG yields the best performance.

\section{Supertagging since 2010}\label{sec:stsc}
Since the publication of the edited volume~\cite{bangalore2010supertagging} on 
supertagging, there has not been much supertagging work in (L)TAG framework. 
However, there has been some ongoing work in the application of 
supertags in CCG framework 
\cite{auli2011comparison,ambati2013using,ambati2014improving}.

\newcite{auli2011comparison} begin with the work of 
\newcite{clark2004importance} who employ supertagging for building a 
wide-coverage CCG parser. However, they point out that the parser derived 
from CCG's lexical categories based supertagger is highly approximate and uses 
an \emph{adaptive} strategy to supply more supertags to the parser if the 
parser fails to arrive at a parse with the current proposed supertags. The 
authors proceed to propose a A* based search algorithm to retrieve the right 
parses in an exact fashion. The authors experiment with an exhaustive CKY parser 
for CCG grammar and, also, with a A* parser and report that the latter performs 
better than the CKY parser coupled with supertagger. 
\newcite{ambati2013using} and \newcite{ambati2014improving} explore ways to 
integrate the CCG supertags into the MALT parser~\cite{nivre2007maltparser} for 
the purpose of improving Hindi and English dependency parsing. Overall, the idea 
of supertagging seems to be active at least in CCG parsing.

In this survey article, I tried to summarize chapters from supertagging 
relevant to statistical parsing by beginning with an introduction to 
TAG and supertagging; followed by survey in learning of supertags; and, its 
application to semantic labeling. The article also summarizes few articles 
falling within the intersection of supertagging and statistical parsing since 
2010.

\bibliographystyle{acl}
\bibliography{myreflnks}

\end{document}